\documentclass[a4paper,fleqn]{cas-dc}

\usepackage[authoryear]{natbib}
\usepackage{lipsum}
\usepackage{mathtools}
\usepackage{subcaption}
\usepackage{float}
\usepackage{url} 

\usepackage[T1]{fontenc}

\newenvironment{function}[1][htb]
{
    \begin{algorithm}[#1]

}
{
    \end{algorithm}
}

\floatstyle{ruled}
\newfloat{algorithm}{tbp}{loa}
\floatname{algorithm}{Algorithm}
\usepackage{algpseudocode} 
\AtBeginEnvironment{thebibliography}{%
  \footnotesize
  \setlength{\itemsep}{0pt}%
  \setlength{\parsep}{0pt}%
  \setlength{\parskip}{0pt}%
}
\def\tsc#1{\csdef{#1}{\textsc{\lowercase{#1}}\xspace}}
\tsc{WGM}
\tsc{QE}

\makeatletter
\def\@thehead{}
\def\@thefoot{}
\makeatother
\ExplSyntaxOn
\bool_gset_true:N \g_stm_nologo_bool
\ExplSyntaxOff

\begin{document}
\hbadness=10000
\hfuzz=1000pt
\vfuzz=1000pt
\let\WriteBookmarks\relax
\def\floatpagepagefraction{1}
\def\textpagefraction{.001}

\shorttitle{Re-examining Granger Causality with CBNs and RCCPs}    

\shortauthors{Sadiq A. Adedayo}  

\title [mode = title]{Re-examining Granger Causality with Causal Bayesian Networks and Reichenbach's Principles}  

\author[1]{S.A. Adedayo}[orcid=0000-0003-2990-142X]

\ead{sadiq.adedayo@univie.ac.at}

\affiliation[1]{organization={Univie Doctoral School of Computer Science (DOCS)},
            addressline={Wahringerstraße 29}, 
            city={Vienna},
            postcode={1090}, 
            state={Vienna},
            country={Austria}}

\begin{abstract}
Granger causality (GC) is widely used to infer directed relationships in time-series data. However, its predictive criterion does not by itself distinguish direct causal effects from dependencies induced by common causes, indirect paths, collider conditioning, or model misspecification. We revisit this limitation by interpreting bivariate and multivariate GC through causal Bayesian networks and Reichenbach's common cause principles. Under explicit graphical assumptions, bivariate GC provides a marginal dependence check, while multivariate GC tests whether the same association persists after conditioning on relevant histories. This view motivates causalised Granger causality (c-GC), which combines the two decisions, and c-GC*, a more conservative variant with a richer conditioning set. We validate both methods on synthetic dynamical systems, established time-series causal discovery benchmarks, Sachs' protein-signalling data, and Lorenz-96 simulations. The results show that the proposed criteria recover plausible causal structure in settings with delayed effects, cycles, bidirectional links, and nonlinear or noisy dynamics. The framework clarifies how GC-style inference can be given a causal interpretation without treating temporal prediction alone as sufficient evidence of causation.
\end{abstract}


{\hfuzz=1000pt\hbadness=10000\maketitle}

\section{Introduction} \label{introduction}

Across scientific fields, machine learning (ML) has made it easier to extract patterns from data and forecast the evolution of complex systems \citep{gil2014amplify, xu2021artificial, bianchini2022artificial}. Good prediction, however, is not causal understanding. A model may forecast a variable accurately without revealing what would change under intervention, or whether an observed association comes from a direct mechanism, an indirect path, a common cause, or sampling effects \citep{pearl2009causality, peters2017elements, scholkopf2021toward}. This distinction is especially sharp in dynamical systems: temporal order is informative, but feedback, delayed responses, nonlinear behaviour, and latent influences can make ordinary associations look causal \citep{cliff2022unifying, runge2018causal, runge2023causal}. Causal structure learning (CSL) therefore begins from a simple but difficult question: when should a dependency in observational data be treated as a causal link \citep{heinze2018causal, scholkopf2022causality}?

Time-series analysis brings this problem into focus, and Granger causality (GC) remains one of its most influential tools. GC uses temporal predictability to infer directed dependence: if the past of one process improves prediction of another, the first process is said to Granger-cause the second \citep{granger1969investigating, granger1988some, shojaie2022granger}. This idea has been useful in econometrics, neuroscience, and other settings where temporal measurements are central \citep{Seth3293, ding2006granger, barnett2014mvgc}. Its strength is also its weakness. Predictive improvement can arise from a direct causal effect, but also from common causes, indirect pathways, collider conditioning, model misspecification, or latent variables \citep{maziarz2015review, stokes2017study, eichler2012causal, grassmann2020new}. GC therefore gives a concrete test for directed temporal association, but it does not by itself settle when that association has a causal interpretation \citep{eichler2010granger, white2010granger, white2011linking}.

This paper revisits that gap through causal Bayesian networks (CBNs) and Reichenbach's common cause principles (RCCPs) \citep{pearl2000models, pearl2016causal, reichenbach1956direction, hofer1999reichenbach}. Rather than treating GC as a standalone causal rule, we read its bivariate and multivariate forms as complementary tests on a time-indexed graphical model. The bivariate test asks whether two variables are associated across time; the multivariate test asks whether that association survives after conditioning on relevant histories. Neither test is sufficient on its own. Together, they suggest a stricter route from temporal dependence to causal interpretation. This observation motivates causalised Granger causality (c-GC) and its conservative variant, c-GC*.

Our contributions are as follows:
\begin{itemize}
    \item Interpret GC through CBNs and RCCPs as a pair of marginal and conditional (in)dependence tests, giving a causal reading to GC-style links under explicit graphical assumptions.
    \item Formalise this interpretation as the c-GC method family, comprising c-GC and the more conservative c-GC*, to reduce false positives caused by confounding, indirect paths, and collider conditioning.
    \item Evaluate both variants on synthetic systems, time-series causal discovery benchmarks, Sachs' protein-signalling data, and Lorenz-96 simulations \citep{sachs2005causal, lorenz1996predictability, runge2019detecting, nauta2019causal, sun2021nts}.
\end{itemize}

The next section supplies the formal vocabulary for this argument by reviewing GC, CBNs, and RCCPs. Section~\ref{sec:meth} then develops the CBN/RCCP interpretation and presents c-GC and c-GC*. The experiments evaluate both methods before the paper concludes.

\section{Background} \label{sec:background}

To make the argument precise, this section reviews the three ingredients used in the method: Granger causality, causal Bayesian networks (CBNs), and Reichenbach's common cause principles (RCCPs).

\subsection{Granger Causality} \label{subsec:granger_causality}

\citet{granger1969investigating, granger1988some} introduced GC as a predictive criterion for causal ordering in time series. GC posits that a time series \(X^{i}\) Granger-causes another time series \(X^{j}\) if including the past values of \(X^{i}\) and \(X^{j}\) in a predictive model improves the prediction of \(X^{j}\) compared to using only the past values of \(X^{j}\). This relationship is formalised as follows (\(\tau\) is a lag parameter):
{\footnotesize
\begin{equation} \label{eqn:GCeqn}
    X^{i} \overset{GC}{\longrightarrow} X^{j} =
    \text{var}\left[X_{t}^{j} - \hat{X}_{t}^{j} \big| X_{0:t-\tau}^{j}\right] > \text{var}\left[X_{t}^{j} - \hat{X}_{t}^{j} \big| X_{0:t-\tau}^{j}, X_{0:t-\tau}^{i}\right].
\end{equation}}
\vspace{-0.6em}
    
An F-test (eqn \eqref{eqn:FTest}) is used to assess the significance of the model. Here, \(RSS_{\text{full}}\) and \(RSS_{\text{red}}\) denote the residual sum of squares for the full model (using past values of both variables) and the reduced model (using only the past values of the predicted variable), respectively. The degrees of freedom are \(p_{2} - p_{1}\) and \(n - p_{2}\), where \(n\) is the number of data points, and \(p_{1}\) and \(p_{2}\) are the number of parameters in the reduced and full models, respectively. Note that \(p_{2} > p_{1}\).
\begin{equation} \label{eqn:FTest}
    F = \frac{(RSS_{\text{red}} - RSS_{\text{full}}) / (p_{2} - p_{1})}{RSS_{\text{full}} / (n - p_{2})}
\end{equation}

Because GC is a predictive criterion rather than an interventionist one, it has been criticised for not representing causality strictly, for relying on linear assumptions in many applications, and for being vulnerable to latent confounders. It is also important to distinguish the general predictive principle of GC
from its classical operational implementation: the definition of GC is based on whether the past of one process improves the prediction of another, whereas the standard formulation typically instantiates this idea using linear autoregressive or vector autoregressive models. Many extensions have therefore been proposed to relax modelling assumptions, handle nonlinear dependencies, incorporate multivariate conditioning, or improve robustness in applied settings \citep{shojaie2022granger, wen2013granger, white2010granger, cadotte2010granger, eichler2010granger, zou2009granger, chen2004analyzing, ancona2004radial, bell1996non, chamberlain1982general, geweke1982measurement}. Within this classical linear time-domain framework, \citet{geweke1982measurement} formalised a variance-ratio measure of linear dependence and feedback between time series, yielding the statistic in equation \eqref{eqn:GCFTest}:
\begin{equation} \label{eqn:GCFTest}
    F_{X^{i} \overset{GC}{\longrightarrow} X^{j}} = \ln \frac{\Big|\text{var}\left(X_{t}^{j} - \hat{X}_{t}^{j^{(r^{i})}} \right)\Big|}{\Big|\text{var}\left(X_{t}^{j} - \hat{X}_{t}^{j^{(f)}} \right)\Big|}.
\end{equation}
\vspace{-0.6em}

Here, \(X_t^j - \hat{X}_t^{j^{(r^i)}}\) denotes the prediction error, or residual, of the restricted model, while \(X_t^j - \hat{X}_t^{j^{(f)}}\) denotes the prediction error, or residual, of the full model. If including \(X_i\) substantially reduces the prediction error variance for \(X_j\), then the ratio is greater than one, and the logarithmic score is positive, providing evidence that \(X_i\) Granger-causes \(X_j\). Related information-theoretic and nonlinear approaches, including intervention-based information flow, transfer entropy, and convergent cross mapping, have also been used to quantify directed dependence or causal influence in complex time-series systems \citep{ay2008information, zheng2020dynamics, wang2026uncovering}.
  
Importantly, temporal precedence$\--$cause variable precedes the effect by a lag parameter \(\tau\)$\--$defining time delay between the cause and effect is critical to the GC framework (i.e., GC does not infer instantaneous connections) \citep{ding2006granger, scholkopf2022causality, eichler2012causal}. There are two GC cases:

\textbf{Bivariate Granger Causality (BVGC)} tests associational (in)dependence between two variables, say \(X^{x}\) and \(X^{y}\), by comparing the prediction variance resulting from full (i.e., including pasts of both \emph{cause} and \emph{effect} variables) compared to reduced (i.e., only the \emph{effect}'s pasts). Equation \eqref{eqn:fullbvgc} formalises BVGC, where \(\tau\) and \(\tau_{max}\) are the link lag and the maximum number of pasts considered, respectively.
    \begin{equation} \label{eqn:fullbvgc}
        X_{t-\tau}^{x} \overset{?}{\longrightarrow} X_{t}^{y} \big| X_{t - (\tau + 1) : t-\tau_{max}}^{x}, X_{t-1 : t-\tau_{max}}^{y}.
    \end{equation}

\textbf{Multivariate Granger Causality (MVGC)} applies the general predictive definition of GC in a conditional setting. Conceptually, \(X^{x}\) Granger-causes \(X^{y}\) if the past of \(X^{x}\) improves the prediction of \(X_{t}^{y}\)
beyond the information already contained in the past of \(X^{y}\) and the past of the remaining variables \(X^{z}\). The classical operational form of MVGC implements this predictive comparison with linear vector autoregressive (VAR) models, although the predictive definition itself is not restricted to linear models. For a candidate lag \(\tau\), this conditional test can be written as
\begin{equation} \label{eqn:fullmvgc}
  X_{t-\tau}^{x} \overset{?}{\longrightarrow} X_{t}^{y} \big|
  X_{t-(\tau + 1):t-\tau_{max}}^{x},
  X_{t-1:t-\tau_{max}}^{y},
  X_{t-1:t-\tau_{max}}^{z}.
\end{equation}

Granger acknowledged that his framework does not reveal true causal structures, but argued that, under certain assumptions, predictability can imply a mechanistic cause-effect relationship \citep{scholkopf2021toward, eichler2012causal, white2011linking, ding2006granger}. This point sets up Section~\ref{sec:meth}, where BVGC and MVGC are recast in the language of CBNs and RCCPs.

Several studies have suggested combining both GC cases to give GC-inferred structures clearer causal interpretations \citep{shojaie2022granger, blinowska2004granger, siggiridou2019evaluation, barrett2010multivariate, stokes2017study, eichler2012causal, ding2006granger}. Section~\ref{sec:meth} gives the CBN/RCCP justification for this combination and presents the resulting algorithms, c-GC and c-GC*.  

\subsection{Causal Bayesian Networks (CBNs)} \label{subsec:cbn}
Causal Bayesian Networks (CBNs) are a class of probabilistic graphical models (PGMs) that explicitly represent causal relationships. CBNs, often synonymous with \emph{causal graphical models}, extend Bayesian networks by incorporating causal semantics. In Bayesian networks, the joint distribution \(P(X_{1}, X_{2}, \dots, X_{n})\) is modeled using the chain rule (eqn \eqref{eqn:BY}):
\begin{equation} \label{eqn:BY}
    P(X_{1}, \dots, X_{n}) = P(X_{1}) \prod_{i} P\left(X_{i} \big| X_{i-1}, \dots, X_{1}\right).
\end{equation}
However, this approach requires an exponential number of parameters. To address this, CBNs factorise the joint distribution based on local dependencies, reducing computational complexity (eqn \eqref{eqn:BY1}). This factorisation relies on the \emph{local Markov assumption}, where each variable depends only on its parents \(Pa(\cdot)\).
\begin{equation} \label{eqn:BY1}
    P(X_{1}, \dots, X_{n}) = \prod_{i} P\left(X_{i} \big| Pa(X_{i})\right).
\end{equation}
Directed graphs \(\mathcal{G}\) are used to denote causal relationships in CBN, where nodes are variables and edges are the causal interactions \citep{pearl2000models}. The joint probability distribution is as in eqn~\ref{eqn:BY1}. While many CSL algorithms assume acyclicity, real-world systems like biological neural networks often exhibit feedback loops, cyclicity and bidirectional interactions. In our approach, we do not enforce any of these characteristics.

\subsection{Reichenbach's Principles}
\label{subsec:reichenbach}

\citet{reichenbach1956direction} introduced the common cause principles, emphasising the role of time in causation. These principles are grounded in the logical notion that no effect precedes its cause in physical systems\footnote{Physical systems are characterised as Markov \citep{weber2009dynamical, ibe2013markov, attal2010markov, van1992stochastic}.}. For a detailed discussion on RCCP, we refer readers to \citep{reichenbach1991direction, hofer1999reichenbach, reichenbach1956direction}. RCCP uses probability measures to define (in)dependence: events \(A\) and \(B\) are dependent if their joint probability exceeds the product of their marginals (equation \eqref{eqn:rccp1}). If neither variable is the direct cause or effect, yet dependence holds, Reichenbach postulated the existence of a common cause occurring earlier, inducing an associational relationship. This common cause can render the pair conditionally independent. RCCP forms the basis of the Causal Markov Condition \citep{neuberg2003causality, pearl2016causal, eells1991probabilistic, salmon1998causality, reichenbach1991direction} and is formalised through the following principles:
\begingroup
\setlength{\jot}{0.25em}
\begin{align}
    P(A, B) &> P(A) \cdot P(B) \label{eqn:rccp1}\\
    P(A, B | C) &= P(A | C) \cdot P(B | C) \label{eqn:rccp2}\\
    P(A, B | C') &= P(A | C') \cdot P(B | C') \label{eqn:rccp3}
\end{align}
\endgroup

Let \(P(A)\) and \(P(B)\) denote the marginal probabilities of variables \(A\) and \(B\), and \(P(A, B)\) their joint probability. Conditional joint probabilities are written as \(P\left(A, B \big| C\right)\). We likened equation \eqref{eqn:rccp1} to BVGC, as it, like BVGC, tests for unconditional dependencies between variable pairs. Equations \eqref{eqn:rccp2} and \eqref{eqn:rccp3} define conditional (in)dependence, where the union of \(C\) and \(C'\) is the set of all other variables in the world. In Reichenbach's framework, a causal link exists if and only if these equations are satisfied.

\section{Method} \label{sec:meth}

After formalising some requirements in section \ref{sec:background} and presenting the limitations of GC, we propose a causal endowment for GC. For simplicity, let $X = \{X^{x}, X^{y}, X^{z}\}$ be a multivariate dataset. $\mathcal{X} = \{X_{t}, \dots, X_{\tau_{max} + 1}\}^{T}$, \(\tau\) is a lag parameter. The constant $1$ was added such that the cause variable at maximum lag $X_{\tau_{max}}$ can have a past state. The analogies on this dataset generalise as we assumed all other variables in the world are summarised into \(X^{z}\), thus satisfying the causal sufficiency assumption.

\subsection{Linking GC to CBNs and RCCPs}

Here, we break down how we connect GC with both CBNs and RCCPs.

\subsubsection{Interpreting GC in CBNs}
\begin{figure}
    \centering
    \begin{subfigure}[b]{0.48\columnwidth}
        \centering
        \includegraphics[scale=0.45]{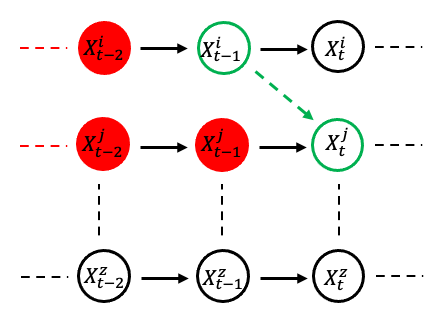}
        \caption{}
        \label{fig:bvgc}
    \end{subfigure}
    \begin{subfigure}[b]{0.48\columnwidth}
        \centering
        \includegraphics[scale=0.45]{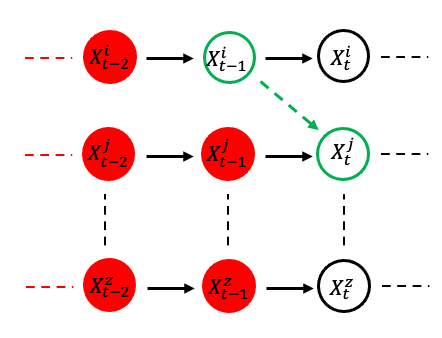}
        \caption{}
        \label{fig:mvgc}
    \end{subfigure}
    \caption{GC interpretation in CBNs as a conditional (in)dependence test. Figure~\ref{fig:bvgc} depicts BVGC conditioning only on the pasts of both the cause \(X_{t-1}^{i}\) and effect \(X_{t}^{j}\) variables. Figure~\ref{fig:mvgc} is the MVGC counterpart, including past states of other variables in the conditioning set. The dashed green arrow indicates the link being tested with \(\tau = 1\). All red-shaded nodes are observed.}
    \label{fig:gc_cbn}
\end{figure}

In CBN lenses, Granger causality can be interpreted as a conditional independence test. Specifically, it tests for statistical (in)dependence between residuals \(\epsilon_{X_{t-\tau}^{i}}\) and \(\epsilon_{X_{t}^{j}}\), obtained by regressing on the conditioning set \(Z\) (containing the past states of the predictors up to the selected lag value (\(\tau\))). This interpretation is depicted in Figure~\ref{fig:gc_cbn}

We grounded our connection between GC and CBNs on the conditioning set ($Z$). In our approach, MVGC uses a well-mapped out $Z$ that fulfils cogent CBN requirements such as back- and front-door criteria. However, the connection arises as we combine the strengths of both MVGC and BVGC. This combination fulfils and enforces all RCCPs, rendering any inferred links \emph{causal}.
\begingroup
\setlength{\jot}{0.25em}
\begin{align}
    \epsilon_{X_{t-\tau}^{i}} &= X_{t-\tau}^{i} \big| Z\\
    \epsilon_{X_{t}^{j}} &= X_{t}^{j} \big| Z
\end{align}
\endgroup

If there exists a statistically significant dependence between the residuals, we can infer \(X_{t-\tau}^{i} \overset{GC}{\longrightarrow} X_{t}^{j}\). For a comprehensive introduction to CBNs, we refer readers to \cite{spirtes2000causation, pearl2000models}.

\subsubsection{Connecting RCCPs to GC}

After making the above clarification, we now factor in Reichenbach's principles and formulate the following three propositions:

\textbf{\textit{Proposition 1:}} \textit{Dependence between any variable pair, irrespective of potential confounders, must hold.}

\textbf{\textit{Illustration:}} BVGC formalised in eqn \eqref{eqn:fullbvgc} tests pairwise associational (in)dependency between two variables without considering the effect of confounders. We likened this to the first RCCP (eqn \eqref{eqn:rccp1}) and hence, reduced eqn~\ref{eqn:fullbvgc} to a simple marginal (in)dependence test (i.e., $X_{t-\tau}^{x} \overset{?}{\not\!\perp\!\!\!\perp} X_{t}^{y}$). This reduction prevents potential spurious dependencies that may arise as a result of observing the past states. Moreso, simplifies the model and improves our runtime. We used this test to identify potential \textit{colliders}\footnote{\textit{Chains}, \textit{forks}, and \textit{colliders} are the three fundamental building blocks of CBNs. We discuss these concepts briefly in relation to our propositions, where we explained the information in Table~\ref{tab:GC-CBN}. For in-depth clarification, we refer readers to  \citep{pearl2016causal, neal2020introduction}}, also referred to as Y- or V-structures, in the dataset. It is easy to see how testing for potential marginal (in)dependence between variable pairs without considering the effect of confounders can only detect (in)dependence, but not sufficient to infer a causal link. 

\textbf{\textit{Proposition 2:}} \textit{Dependence between any variable pair, considering all other variables must hold.}

\textbf{\textit{Illustration:}} MVGC tests pairwise (in)dependency of two variables given all other variables as described in eqn \eqref{eqn:fullmvgc}. Let \(Z_{MVGC}\) be the conditioning set in eqn \eqref{eqn:fullmvgc} for the sake of brevity from here onwards. Concretely, this proposition focuses on testing (in)dependence in both chains and forks and, importantly, upholds eqns \eqref{eqn:rccp2} and \eqref{eqn:rccp3}. 

\textbf{\textit{Proposition 3:}} \textit{Combining\footnote{Combination operation using the logical-and (\(\land\))} propositions \(1\) and \(2\) enforces all RCCPs and reveals only true causal links.}

\textit{Illustration:} Taking the \emph{logical-and} of the binarised inference from both unconditional and conditional dependence tests satisfactorily enforces all of Reichenbach's principles. BVGC, on its own, will infer more connections as we do not condition on possible confounders, which leads to inferring spurious links. On the other hand, MVGC would reveal that some BVGC inferred connections are spurious connections, thus weeding them out. However, MVGC would also infer spurious connections when colliders are conditioned on; in turn, BVGC would have detected and failed to infer. Hence, the \textit{logical-and} operation on the results of both BVGC and MVGC gives rise to a true causal connectivity matrix.

It has been established that regressing in the true causal direction (from cause to effect) typically results in residuals that are independent of the predictor. In contrast, regressing from effect to cause results in dependent residuals on the predictor. This realisation enables the identification of causal directions in non-Gaussian data but not in Gaussian data \citep{shimizu2006linear, Hyvarinen2013, peters2017elements, hoyer2008nonlinear, monti2020causal, mooij2016distinguishing, pearl2016causal}. We take advantage of this knowledge to transition from simple dependence to inferring directed causal links when we enforce both propositions $1$ and $2$. This is important as mere dependence is symmetric \citep{bontempi2015dependency, eichler2012causal, neuberg2003causality, jaynes1957information}.

Table~\ref{tab:GC-CBN} summarises our illustrations via logical examples using the three CBN building blocks, with each example respecting time causality. True DAG column indicates the assumed ground truth, and red-coloured variables are observed. Green statements are the wrongly inferred (i.e., false positives). Blue texts, on the other hand, are correctly inferred links resulting from the combination of BVGC and MVGC.

\begin{enumerate}
    \item \textbf{\emph{Chains}}: Sequential causation such as $X_{t-2}^1 \rightarrow X_{t-1}^3 \rightarrow X_t^2$. BVGC would infer a spurious link $\color{teal}X_{t-2}^{1} \not\!\perp\!\!\!\perp X_{t}^{2}\color{black}$ due to associational flow of information through \(X_{t-1}^{3}\). On the other hand, MVGC which conditions on $X_{t-1}^3$ will correctly reveal independence \(\left(X_t^2 \!\perp\!\!\!\perp X_{t-2}^1 \big| X_{t-1}^3\right)\).

    \item \textbf{\emph{Forks}}: Also referred to as common cause structure e.g., $X_t^2 \leftarrow X_{t-1}^3 \rightarrow X_t^1$. Like chains, they exhibit the same Markov equivalence class (MEC), i.e., they share the same set of conditional independence statements. BVGC will wrongfully infer $\color{teal}X_t^1 \not\!\perp\!\!\!\perp X_t^2\color{black}$, but MVGC, conditioning on $X_{t-1}^3$, will correctly reveal they are in fact independent \(\left(X_t^2 \!\perp\!\!\!\perp X_{t-2}^1 \big| X_{t-1}^3\right)\).

    \item \textbf{\emph{Collider}:} Popularly known as V- or Y-structures such as $X_{t-1}^1 \rightarrow X_t^3 \leftarrow X_{t-1}^2$ are a special case and do not share MEC. In this scenario, BVGC will correctly infer $X_{t-1}^1$ and $X_{t-1}^2$ are independent (i.e., $X_{t-1}^1 \!\perp\!\!\!\perp X_{t-1}^2$), but MVGC, by conditioning on $X_t^3$ will induces associational dependence as $\textcolor{teal}{X_{t-1}^{1} \not\!\perp\!\!\!\perp X_{t-1}^{2} \big| X_t^{3}}$.
\end{enumerate}

In summary, \textit{MVGC} between any independent variable pair that are common causes of another variable would reveal a spurious causal link (false positive) resulting from associational dependence induced by conditioning on the common cause. In contrast, \textit{BVGC} should detect no link because the two variables are independent. Combining the power of both GC cases with the \emph{AND} operator enhances both inferences and reveals true causal links.

\begin{table}[ht]
    \centering
    \caption{Illustrating BVGC and MVGC expected results on example data. False positives are depicted in green, blue are true positives, and red are conditioned variables.}
    \label{tab:GC-CBN}
    \small 
    \resizebox{\columnwidth}{!}{ 
        \begin{tabular}{|c|c|c|c|}
            \hline\hline
            \textbf{True DAG} & \textbf{BVGC} & \textbf{MVGC} & \textbf{Inferences} \\[0.3cm]
            \hline\hline
            \textbf{Chain} & & & \\[0.3cm]
            \makecell[c]{$X_{t-2}^{1} \longrightarrow \color{red}X_{t-1}^{3}\color{black} \longrightarrow X_{t}^{2}$} & 
            \makecell[c]{$\color{teal}X_{t-2}^{1} \not\!\perp\!\!\!\perp X_{t}^{2} \color{black}$} & 
            \makecell[c]{$X_{t-2}^{1} \perp\!\!\!\perp X_{t}^{2} \mid X_{t-1}^{3}$} & \\[0.3cm]
            & 
            \makecell[c]{$X_{t-1}^{3} \not\!\perp\!\!\!\perp X_{t}^{2}$} & 
            \makecell[c]{$X_{t-1}^{3} \not\!\perp\!\!\!\perp X_{t}^{2} \mid X_{t-2}^{1}$} & 
            \makecell[c]{$\color{blue}X_{t-1}^{3} \longrightarrow X_{t}^{2}\color{black}$} \\[0.3cm]
            & 
            \makecell[c]{$X_{t-2}^{1} \not\!\perp\!\!\!\perp X_{t-1}^{3}$} & 
            \makecell[c]{$X_{t-2}^{1} \not\!\perp\!\!\!\perp X_{t-1}^{3} \mid X_{t}^{2}$} & 
            \makecell[c]{$\color{blue}X_{t-2}^{1} \longrightarrow X_{t-1}^{3}\color{black}$} \\[0.3cm]
            \hline
            \textbf{Forks} & & & \\[0.3cm]
            \makecell[c]{$X_{t}^{1} \longleftarrow \color{red}X_{t-1}^{3}\color{black} \longrightarrow X_{t}^{2}$} & 
            \makecell[c]{$\color{teal}X_{t}^{2} \not\!\perp\!\!\!\perp X_{t}^{1}\color{black}$} & 
            \makecell[c]{$X_{t}^{2} \perp\!\!\!\perp X_{t}^{1} \mid X_{t-1}^{3}$} & \\[0.3cm]
            & 
            \makecell[c]{$X_{t-1}^{3} \not\!\perp\!\!\!\perp X_{t}^{2}$} & 
            \makecell[c]{$X_{t-1}^{3} \not\!\perp\!\!\!\perp X_{t}^{2} \mid X_{t}^{1}$} & 
            \makecell[c]{$\color{blue}X_{t-1}^{3} \longrightarrow X_{t}^{2}\color{black}$} \\[0.3cm]
            & 
            \makecell[c]{$X_{t-1}^{3} \not\!\perp\!\!\!\perp X_{t}^{1}$} & 
            \makecell[c]{$X_{t-1}^{3} \not\!\perp\!\!\!\perp X_{t}^{1} \mid X_{t}^{2}$} & 
            \makecell[c]{$\color{blue}X_{t-1}^{3} \longrightarrow X_{t}^{1}\color{black}$} \\[0.3cm]
            \hline
            \textbf{Colliders} & & & \\[0.3cm]
            \makecell[c]{$X_{t-1}^{1} \longrightarrow \color{red}X_{t}^{3}\color{black} \longleftarrow X_{t-1}^{2}$} & 
            \makecell[c]{$X_{t-1}^{1} \perp\!\!\!\perp X_{t-1}^{2}$} & 
            \makecell[c]{$\color{teal}X_{t-1}^{1} \not\!\perp\!\!\!\perp X_{t-1}^{2} \mid X_{t}^{3}\color{black}$} & \\[0.3cm]
            & 
            \makecell[c]{$X_{t-1}^{2} \not\!\perp\!\!\!\perp X_{t}^{3}$} & 
            \makecell[c]{$X_{t-1}^{2} \not\!\perp\!\!\!\perp X_{t}^{3} \mid X_{t-1}^{1}$} & 
            \makecell[c]{$\color{blue}X_{t-1}^{2} \longrightarrow X_{t}^{3}\color{black}$} \\[0.3cm]
            & 
            \makecell[c]{$X_{t-1}^{1} \not\!\perp\!\!\!\perp X_{t}^{3}$} & 
            \makecell[c]{$X_{t-1}^{1} \not\!\perp\!\!\!\perp X_{t}^{3} \mid X_{t-1}^{2}$} & 
            \makecell[c]{$\color{blue}X_{t-1}^{1} \longrightarrow X_{t}^{3}\color{black}$} \\[0.3cm]
            \hline
        \end{tabular}
    }
\end{table}

Dependence can be tested by various means, including, but not limited to, Pearson's correlation, Hilbert-Schmidt Independence Criterion (HSIC), normalised mutual information, and others \citep{bontempi2015dependency, granger2004dependence, wu2010new, gretton2007kernel, nalewajski2011elements, cover1999elements}. In this work, we adopted permutation-based Pearson's correlation due to its simplicity and non-parametric nature.

\subsection{The Algorithm (c-GC and c-GC*)}

Here, we formalise the above-discussed propositions into an algorithm (c-GC). Let \(X\) be a multivariate time series dataset with \(n\) variables. The pseudocode \emph{c-GC()} formalises our approach, ensuring the following assumptions: 
\begin{itemize}
    \item No connections exist outside of chosen $\tau_{max}$.
    \item Causal Markov Assumption: All variables are conditionally independent of their non-descendants given their parents. 
    \item Causal sufficiency: all variables are contained in \(\mathcal{X}\) emphasizing the absence of latent variables.
    \item Faithfulness: All conditional (in)dependence in graph \(\mathcal{G}\) are implied by its causal structure.
\end{itemize}

Before formalising \emph{c-GC()}, we introduce some parameters and helper functions.

\subsubsection{Algorithm parameters}

\begin{itemize}
    \item significance levels ($\alpha$, $\beta$): Hyperparameters set by the user. We recommend conservative values ($\alpha = 0.01$ and $\beta = 0.001$) for \textit{BVGC()} and \textit{MVGC()} tests, respectively, as experiments showed better performance with stricter $\beta$.
    
    \item lag ($\tau$): Time delay between cause and effect variables. A big maximum lag ($\tau_{max}$) increases computational runtime.
    
    \item number of permutations ($N$): By default, $N = 1000$ for $T \leq 5000$. Larger $N$ improves accuracy but slows computation. The p-value is computed as:
    \begin{equation} \label{eqn:pVal}
        \text{p-value} = \frac{T_{\text{perm}} \geq T_{\text{obs}}}{N} 
    \end{equation}

    where $T_{perm}$ and $T_{obs}$ are the permutation statistic and the observed test statistic, respectively.
    
\end{itemize}

\subsubsection{Conditioning set (Z)} \label{sec:condSets}
Following the GC framework, our method relies on \(Z\), which is constructed using the variables' pasts. \(Z\) satisfies the d-separation and back-door criteria, ensuring completeness. These criteria are met in three scenarios.

\begin{figure}
    \centering
    \begin{subfigure}[b]{0.32\columnwidth}
        \centering
        \includegraphics[scale=0.34]{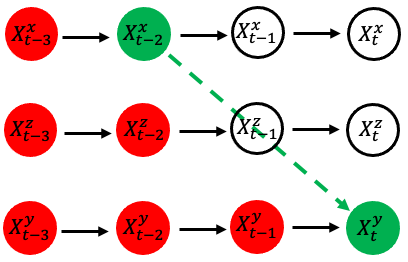}
        \caption{}
        \label{fig:my_GC_CondSet1}
    \end{subfigure}
    \begin{subfigure}[b]{0.32\columnwidth}
        \centering
        \includegraphics[scale=0.34]{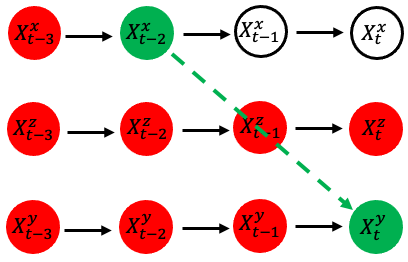}
        \caption{}
        \label{fig:my_GC_CondSet2}
    \end{subfigure}
    \begin{subfigure}[b]{0.32\columnwidth}
        \centering
        \includegraphics[scale=0.34]{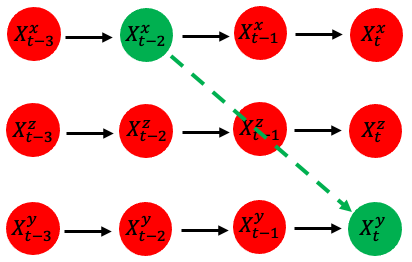}
        \caption{}
        \label{fig:my_GC_CondSet3}
    \end{subfigure}
    \caption{%
        Conditioning set scenarios. The dotted green arrow indicates the link being tested ($\tau = 2$ for clarity). Variables in $Z$ are shaded in red. c-GC and c-GC* use scenarios (b) and (c), respectively.}
    \label{fig:CondSet}
\end{figure}

\begin{enumerate}
    \item Depicted in Figure~\ref{fig:my_GC_CondSet1}; $Z$ contains the past states of all variables but excludes all states of $X^{z}$ before the cause variable $X^{x}_{t-\tau}$.
    \begin{equation}
        Z = \{\mathcal{X}\backslash[X^{x}_{t:t-\tau}, X^{y}_{t}, X^{z}_{t:t-\tau}] \}
    \end{equation}
    \item Depicted in Figure~\ref{fig:my_GC_CondSet2}; Unlike above, all states of $X^{z}$ are included in the conditioning set.
    \begin{equation}
        Z = \{\mathcal{X}\backslash[X^{x}_{t:t-\tau}, X^{y}_{t}] \}
    \end{equation}
    \item Depicted in Figure~\ref{fig:my_GC_CondSet3}. This scenario contains all variable states except the pair of interest.    
    \begin{equation}
        Z = \{\mathcal{X}\backslash[X^{x}_{t-\tau}, X^{y}_{t}] \}.
    \end{equation}
\end{enumerate}
Scenarios 1 and 2 yield similar results, but we chose Scenario 1 as the conditioning set ($Z_{MVGC}$) for c-GC. We extended this to Scenario 3 (called c-GC*), which produces comparable results while being the most conservative approach. While c-GC* conditions on the \textit{cause's} future states, the theoretical justification for this remains unclear. Arguably, future states shouldn't exist; however, this is possible as a result of our data manipulation as part of our causal discovery process.  Function $get\_conditioning\_set()$ describes how we extract $Z_{MVGC}$:
\begin{function}[ht!]
    \caption{get\_conditioning\_set()}
    \label{func:get_conditioning_set}
    \begin{algorithmic}[1]
        \Require $\mathcal{X}, i, j, \text{method}==\texttt{cgc}$
        \For{$\forall$ $i$ and $j$ pair}
            \State $Z \gets$ \text{nd.array}
            \If{$\text{not method}$}
                \State $Z \gets \mathcal{X}\setminus[X^{x}_{t-\tau}, X^{y}_{t}]$
            \Else
                \State $Z \gets \mathcal{X}\setminus[X^{x}_{t:t-\tau}, X^{y}_{t}]$
            \EndIf
        \EndFor
        \Ensure $Z$
    \end{algorithmic}
\end{function}

\subsubsection{Inferred connectivity matrix \texorpdfstring{($\hat{A}$)}{(A-hat)}}

Our extended inferred connectivity matrix $\hat{A}_{ext}$ encompasses submatrices encoding the connectivity at different lags ($\tau$). It is a square matrix of shape $[m, m]$, where $m = \mathcal{X}.\text{shape[0]}$, which is in line with the extended summary graph in \citep{assaad2022discovery}. In a strategy to reduce the runtime of our algorithm, we only populate submatrices that encodes the lagged \textit{cause} ($X_{t-\tau}^{i}$) to a fixed present state \textit{effect} ($X_{t}^{j}$) variables (i.e., $X_{t-\tau}^{i} \longrightarrow X_{t}^{j}$). The resulting shape of $\hat{A}_{ext}$ is $[m, n]$, where $n$ is the number of variables and $m = n * (\tau_{max} + 1)$. We make inferences only on selected submatrices up to the maximum allowable lag ($\tau_{max}$), which would correspond to the GC order. We aggregate the inferences from all submatrices corresponding to each $\tau \in \{0, 1, \cdots, \tau_{max}\}$ with $\tau = 0$ denoting the instantaneous links submatrix. \emph{get\_connectivity\_mat()} extracts aggregated inferred connectivity matrix ($\hat{A}$).

\begin{function}[htbp]
    \caption{get\_connectivity\_mat()}
    \label{func:get_connectivity_matrix}
    \begin{algorithmic}[1]
        \Require $\hat{A}_{ext}, \tau_{max}$
        \State Split $\hat{A}_{ext}$ into $\hat{A}_{\tau}$ \Comment{$\tau \in \{0,\dots,\tau_{max}+1\}$}
        \State $\hat{A} \gets \hat{A}_{\tau = 1}$
        \For{$i = 1 \to \tau_{max} + 1$}
            \State $\hat{A} \gets \text{logical\_or}(\hat{A}, \hat{A}_{i})$
        \EndFor
        \State Optional: Remove self-connections \Comment{Optional step}
        \Ensure $\hat{A}$
    \end{algorithmic}
\end{function}

After all the parameters and helper functions have been introduced, we bring them all together and formalise c-GC. Recall that c-GC* is implemented the same, but differs from c-GC based on the $Z_{MVGC}$ used. The implementation can be found at https://github.com/adedayoas91/causalized-GC.

\begin{function}[!ht]
    \caption{c-GC()}
    \label{func:c-GC}
    \begin{algorithmic}[1]
        \Require $X, \tau_{max}, N,$ method
        \State $\mathcal{X} \gets \text{shiftdata}(X, \tau_{max})$
        \State $m, n \gets \mathcal{X}.\text{shape}[0], X.\text{shape}[0]$ \Comment{$m := n * (\tau_{max}+1$)}\\
        \Comment{\# Initialize $A_{BVGC}$ and $A_{MVGC}$}
        \For{$i \to m$}
            \For{$j \to n$}\\
                \State $A^{i,j}_{BVGC} \gets BVGC(X^{i}, X^{j})$ \\ 
                \Comment{Extract $Z_{MVGC}$ for MVGC()}
                \State $Z_{MVGC} \gets \text{get\_conditioning\_set}($
                \Statex \hspace{\algorithmicindent}$\mathcal{X}, i, j, \text{method})$
                \State $A^{i,j}_{MVGC} \gets MVGC\left(X^{i}, X^{j} \mid Z_{MVGC}\right)$
            \EndFor
        \EndFor
        \State $\hat{A}_{ext} \gets A_{BVGC} \land A_{MVGC}$
        \State $\hat{A} \gets \text{get\_connectivity\_mat}(\hat{A}_{ext}, \tau_{max})$
        \Ensure $\hat{A}$
    \end{algorithmic}
\end{function}

\subsubsection{Computational complexity}

Let \(n\) be the number of variables, \(T' = T-\tau_{\max}\) the shifted sample length, \(m=n(\tau_{\max}+1)\) the number of lag-expanded variables, \(N\) the number of permutations, and \(q\) the conditioning-set size. The algorithm tests \(mn\) lagged source--target pairs. The unconditional \emph{BVGC()} pass costs \(\mathcal{O}(mnNT')\), while the conditional \emph{MVGC()} pass dominates because each pair requires dense least-squares residualisation. The overall fitting cost is
\(\mathcal{O}\!\left(m^{2}T' + mn\left(NT' + T'q^{2} + q^{3}\right)\right).\) In the worst case \(q=\mathcal{O}(m)\). Thus, c-GC and c-GC* have the same asymptotic complexity, although c-GC* is typically slower because it uses the largest conditioning set. The final aggregation costs only \(\mathcal{O}(\tau_{\max}n^{2})\), and memory usage is \(\mathcal{O}(mT' + m^{2})\).

\begin{NoHyper}
\section{Experiments}

Here, validation and experimentation results on simulations and observational data are reported. Section~\ref{sec:simu} presents validation of both c-GC and c-GC* and comparison against selected CSL algorithms in section \ref{sec:simu}, Sach's protein dataset in section \ref{sec:sachs} and Lorenz-96 dataset section \ref{sec:lorenz-96}.

\subsection{Synthetic datasets} \label{sec:simu}

We simulated data using an autoregressive model with a known ground-truth (GT) matrix $A_{\tau}$, where $\tau \in \{1, 2\}$ determines the connections and their lags. Equation \ref{eqn:simARModel} formalises our data simulation, with $n = 30$ time series and $T = 5000$ samples. Our scope focuses on aggregate inference \(\hat{A}\) and not the inferred connections' lags.
\begingroup
\setlength{\jot}{0.15em}
\begin{align}
    X_{t+1}^{i} &= X_{t}^{i} + \sum_{X^{j} \in Pa(X^{i})} A^{i,j}_{\tau} X_{t}^{j} + \epsilon^{i}_{t}
    \label{eqn:simARModel}\\
    \epsilon &= \epsilon_{\text{in}} \sim \mathcal{N}(\mu, \sigma) + \epsilon_{\text{ct}}
    \label{eqn:addNoise}
\end{align}
\endgroup

\begingroup\hfuzz=1000pt
\begin{figure}
    \centering
    \begin{subfigure}[b]{0.48\columnwidth}
        \centering
        \includegraphics[scale=0.45]{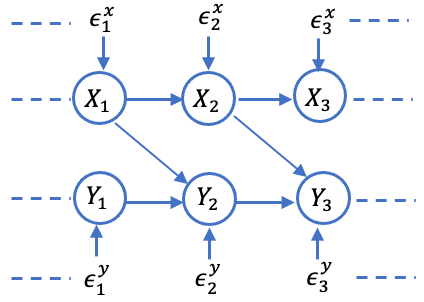}
        \caption{}
        \label{fig:indepTime}
    \end{subfigure}
    \begin{subfigure}[b]{0.48\columnwidth}
        \centering
        \includegraphics[scale=0.45]{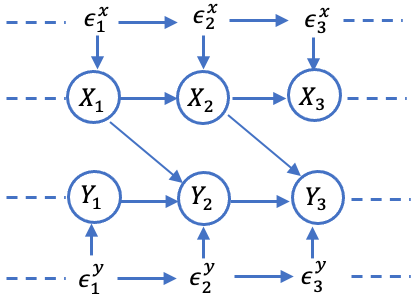}
        \caption{}
        \label{fig:CorrTime}
    \end{subfigure}
    \caption{%
    Additive noise scenarios. (a) independent-time ($\epsilon_{in}$), (b) correlated-time ($\epsilon_{ct}$) noise scenarios.}
    \label{fig:DataScenarios}
\end{figure}
\endgroup 

where $A$ is the ground-truth connectivity matrix, which encodes links at lags $\tau$. $A$ combines all ${A}_{\tau}: \forall \tau \in \{0, 1, \dots, \tau_{max}\}$, can contain cycles and bi-directional links to simulate the connectivity structure of complex systems such as the biological brain. $\epsilon$ indicates a time-correlated, non-Gaussian additive noise imposed on each variable in the data, which controls chaos in our experiment. These intricacies are introduced to show the robustness of our method. Additive noises $\epsilon_{in}$ and $\epsilon_{ct}$ are depicted in Figure~\ref{fig:DataScenarios}

\textbf{Results}: We have data scenarios with and without correlated-time noise $\epsilon_{ct}$. Also, there are two data cases: first with only one inbuilt lag (i.e., $\tau = 1$) and second, with varying time lag $\tau = {1, 2}$. Figure \ref{fig:data_example} shows the obtained results for a linear and minimal chaotic system when $\epsilon_{ct} = 0$. Both c-GC and c-GC* obtained exceptional and similar metrics. c-GC* (see Fig~\ref{fig:res_cGC2}) inferred 1 FP (False Positive) in addition to the 3 FN (False Negative) compared to c-GC (see Fig~\ref{fig:res_cGC1}) which correctly inferred all links (True positives (TP), colour-coded yellow) in figure \ref{fig:res_cGC1} but with three (3) false negatives (FN). 

On the other scenario with $\epsilon_{ct}$, which proves more challenging, we compared inferences from other CSL algorithms alongside other experimental terrains and reported computed metrics such as Accuracy, Precision, Recall, False Positive Rate (FPR), Structural Hamming Distances (SHD), Balanced Accuracy (BA) and F1 scores in Table~\ref{tab:comparisonRes}.

\begin{figure}
    \centering
    \begin{subfigure}[b]{0.275\columnwidth}
        \centering
        \includegraphics[width=\linewidth]{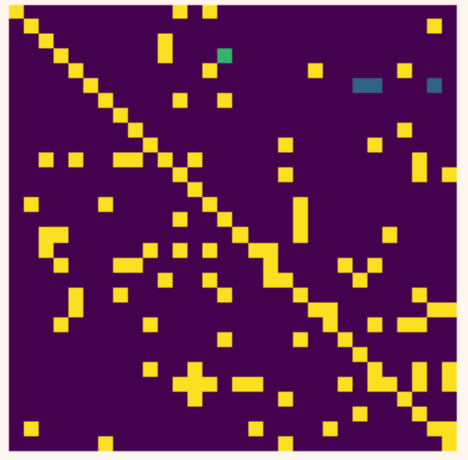} 
        \caption{\(\hat{A}_{c-GC*}\)}
        \label{fig:res_cGC2}
    \end{subfigure}
    \hfill
    \begin{subfigure}[b]{0.275\columnwidth}
        \centering
        \includegraphics[width=\linewidth]{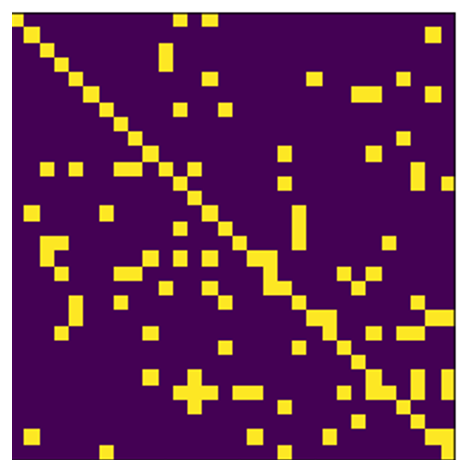}
        \caption{\(A_{GT}\)}
        \label{fig:A_mat}
    \end{subfigure}
    \hfill
    \begin{subfigure}[b]{0.355\columnwidth}
        \centering
        \includegraphics[width=\linewidth]{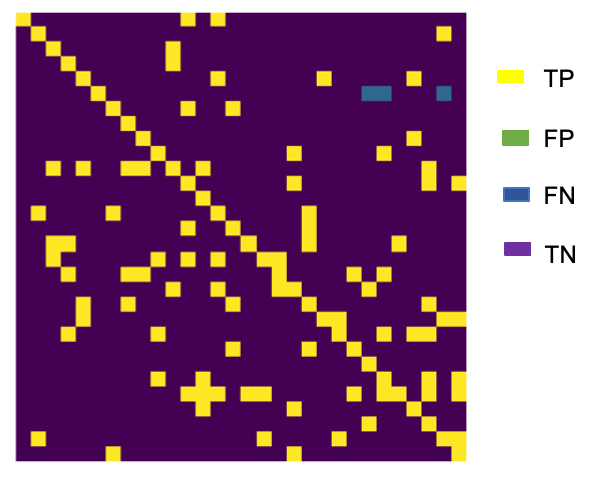} 
        \caption{\(\hat{A}_{c-GC}\)}
        \label{fig:res_cGC1}
    \end{subfigure}
    \caption{Side-by-side view of ground truth matrix \(A\) and inferred matrices \(\hat{A}\). TP, FP, FN, and TN represent true positives, false positives, true negatives, and false negatives, respectively.}
    \label{fig:data_example}
\end{figure}

\subsubsection{Comparison with common CSL algorithm}

This section compares both c-Gc and c-GC* inferences against selected CSL algorithms, including TCDF \citep{nauta2019causal}, NTS-NOTEARS \citep{sun2021nts}, and Tigramite \citep{runge2019detecting}. We did this comparison on various experimental terrains presented by each algorithm to show the robustness of our method. To ensure a fair and rigorous comparison, the algorithms were chosen based on their reputations in the field and their experimental setups. Evaluations were performed using the same datasets on which these algorithms were originally validated and vice versa. The metrics are derived from each method's final connectivity matrices \(\hat{A}\), without regard to the specific time lags at which connections were inferred.  

\begin{table*}
\caption{Performance comparison with selected algorithms against c-GC and c-GC*.}
\centering
\begin{tabular}{l l c c c c c c c}
    \hline\hline
    \textbf{Data} & \textbf{Algorithm} & \textbf{Accuracy} & \textbf{Precision} & \textbf{Recall} & \textbf{FPR} & \textbf{SHD} & \textbf{BA} & \textbf{F1-score} \\
    \hline
    \begin{tabular}[c]{@{}c@{}c@{}c@{}}Simulation \\ with single \\ lag (i.e., $A_{\tau}$, \\ for $\tau = 1$)\end{tabular} 
    & \begin{tabular}[c]{@{}l@{}c@{}c@{}} c-GC* \\ c-GC \\ NTS-NOTEARS \\ TCDF \\ Tigramite(PCMCI) \\ Tigramite(FullCI)\end{tabular} 
    & \begin{tabular}[c]{@{}c@{}@{}c@{}} \textbf{0.9967} \\ 0.9922 \\ 0.8467 \\ 0.8756 \\ 0.4878 \\ 0.8778 \end{tabular} 
    & \begin{tabular}[c]{@{}c@{}c@{}c@{}} \textbf{0.9912} \\ 0.9426 \\ 0.1290 \\ 0.0348 \\ 0.1997 \\ 0.5111 \end{tabular} 
    & \begin{tabular}[c]{@{}c@{}c@{}c@{}} 0.9391 \\ \textbf{1} \\ 0.0348 \\ 0.800 \\ \textbf{1} \\ \textbf{1} \end{tabular} 
    & \begin{tabular}[c]{@{}c@{}c@{}c@{}} \textbf{0.0013} \\ 0.0089 \\ 0.0344 \\ 0.1240 \\ 0.5873 \\ 0.1401 \end{tabular}
    & \begin{tabular}[c]{@{}c@{}c@{}c@{}} 71.94 \\ 77.83 \\ - \\ - \\ - \\ - \end{tabular}
    & \begin{tabular}[c]{@{}c@{}c@{}c@{}} \textbf{0.9907} \\ 0.9955 \\ 0.5002 \\ 0.8380 \\ 0.7063 \\ 0.9299 \end{tabular}
    & \begin{tabular}[c]{@{}c@{}c@{}c@{}} \textbf{0.9869} \\ 0.9705 \\ 0.0548 \\ 0.0667 \\ 0.3329 \\ 0.6765 \end{tabular} \\
    \midrule

    \begin{tabular}[c]{@{}l@{}}Simulation \\ with multiple \\ lags (i.e., $A_{\tau}$, \\ for $\tau \in \{1, 2\}$)\end{tabular} 
    & \begin{tabular}[c]{@{}l@{}}c-GC* \\ c-GC \\ NTS-NOTEARS \\ TCDF \\ Tigramite(PCMCI) \\ Tigramite(FullCI)\end{tabular} 
    & \begin{tabular}[c]{@{}c@{}} \textbf{0.9944} \\ \textbf{0.9944} \\ 0.8167 \\ 0.8778 \\ 0.9422 \\ 0.9600 \end{tabular}  
    & \begin{tabular}[c]{@{}c@{}} \textbf{0.9741} \\ \textbf{0.9741} \\ 0.0833 \\ 0.0609 \\ 0.6886 \\ 0.7616 \end{tabular} 
    & \begin{tabular}[c]{@{}c@{}} \textbf{0.9826} \\ \textbf{0.9826} \\ 0.0434 \\ 0.7778 \\ \textbf{1} \\ \textbf{1} \end{tabular} 
    & \begin{tabular}[c]{@{}c@{}} \textbf{0.0038} \\ \textbf{0.0038} \\ 0.0700 \\ 0.1212 \\ 0.0662 \\ 0.0459 \end{tabular}
    & \begin{tabular}[c]{@{}c@{}} 79.19 \\ 79.19 \\ - \\ - \\ - \\ - \end{tabular}
    & \begin{tabular}[c]{@{}c@{}} \textbf{0.9894} \\ \textbf{0.9894} \\ 0.4867 \\ 0.8283 \\ 0.9669 \\ 0.9771 \end{tabular}
    & \begin{tabular}[c]{@{}c@{}c@{}c@{}} \textbf{0.9784} \\ \textbf{0.9784} \\ 0.0571 \\ 0.1129 \\ 0.8156 \\ 0.8647 \end{tabular} \\
    \midrule

    \begin{tabular}[c]{@{}l@{}}TCDF \\ simulation data\end{tabular} 
    & \begin{tabular}[c]{@{}l@{}}c-GC* \\ c-GC \\ NTS-NOTEARS \\ TCDF \\ Tigramite(PCMCI) \\ Tigramite(FullCI) \end{tabular} 
    & \begin{tabular}[c]{@{}c@{}} 1 \\ 1 \\ 1 \\ 1 \\ 0.8750 \\ 0.9375 \end{tabular} 
    & \begin{tabular}[c]{@{}c@{}} 1 \\ 1 \\ 1 \\ 1 \\ 0.7500 \\ 0.8571 \end{tabular} 
    & \begin{tabular}[c]{@{}c@{}} 1 \\ 1 \\ 1 \\ 1 \\ 1 \\ 1 \end{tabular} 
    & \begin{tabular}[c]{@{}c@{}} 0 \\ 0 \\ 0 \\ 0 \\ 0.2000 \\ 0.1000 \end{tabular}
    & \begin{tabular}[c]{@{}c@{}} 0 \\ 0 \\ 0 \\ 0 \\ - \\ - \end{tabular}
    & \begin{tabular}[c]{@{}c@{}} 1 \\ 1 \\ 1 \\ 1 \\ 0.9000 \\ 0.9500 \end{tabular} 
    & \begin{tabular}[c]{@{}c@{}} 1 \\ 1 \\ 1 \\ 1 \\ 0.8571 \\ 0.9231 \end{tabular} \\
    \midrule

    \begin{tabular}[c]{@{}l@{}}NTS-NOTEARS \\ simulation data\end{tabular} 
    & \begin{tabular}[c]{@{}l@{}}c-GC* \\ c-GC \\ NTS-NOTEARS \\ TCDF \\ Tigramite(PCMCI) \\ Tigramite(FullCI)\end{tabular} 
    & \begin{tabular}[c]{@{}c@{}} 0.6800 \\ 0.6800 \\ \textbf{0.9600} \\ 0.4800 \\ 0.4800 \\ 0.4800 \end{tabular}  
    & \begin{tabular}[c]{@{}c@{}} 0.6111 \\ 0.6111 \\ \textbf{1} \\ 0.3333 \\ 0.4782 \\ 0.4783 \end{tabular} 
    & \begin{tabular}[c]{@{}c@{}} \textbf{0.9200} \\ \textbf{0.9200} \\ 0.9167 \\ 0.4444 \\ 0.9167 \\ 0.9167 \end{tabular} 
    & \begin{tabular}[c]{@{}c@{}} 0.5385 \\ 0.5385 \\ \textbf{0} \\ 0.5000 \\ 0.9231 \\ 0.9231 \end{tabular}
    & \begin{tabular}[c]{@{}c@{}} 7.1730 \\ 7.1730 \\ - \\ - \\ - \\ - \end{tabular}
    & \begin{tabular}[c]{@{}c@{}} 0.6891 \\ 0.6891 \\ \textbf{0.9583} \\ 0.4722 \\ 0.4968 \\ 0.4968 \end{tabular}
    & \begin{tabular}[c]{@{}c@{}} 0.7333 \\ 0.7333 \\ \textbf{0.9565} \\ 0.3809 \\ 0.6285 \\ 0.6285 \end{tabular} \\
    \midrule

    \begin{tabular}[c]{@{}l@{}}Tigramite \\ simulation data\end{tabular} 
    & \begin{tabular}[c]{@{}l@{}}c-GC* \\ c-GC \\ NTS-NOTEARS \\ TCDF \\ Tigramite(PCMCI) \\ Tigramite(FullCI)\end{tabular} 
    & \begin{tabular}[c]{@{}c@{}} 1 \\ 1 \\ 1 \\ 0.7500 \\ 1 \\ 1 \end{tabular}  
    & \begin{tabular}[c]{@{}c@{}} 1 \\ 1 \\ 1 \\ 0.5000 \\ 1 \\ 1 \end{tabular} 
    & \begin{tabular}[c]{@{}c@{}} 1 \\ 1 \\ 1 \\ 1 \\ 1 \\ 1 \end{tabular} 
    & \begin{tabular}[c]{@{}c@{}} 0 \\ 0 \\ 0 \\ 0.3333 \\ 0 \\ 0 \end{tabular}
    & \begin{tabular}[c]{@{}c@{}} 0 \\ 0 \\ 0 \\ - \\ 0 \\ 0 \end{tabular}
    & \begin{tabular}[c]{@{}c@{}} 1 \\ 1 \\ 1 \\ - \\ 1 \\ 1 \end{tabular}
    & \begin{tabular}[c]{@{}c@{}} 1 \\ 1 \\ 1 \\ 0.6666 \\ 1 \\ 1 \end{tabular} \\
    \bottomrule
    \label{tab:comparisonRes}
\end{tabular}
\end{table*}

The performance metrics of each method across various datasets are tabulated in Table~\ref{tab:comparisonRes}. Our methods achieved satisfactory results in most simulation scenarios, demonstrating their robustness and effectiveness. However, this is not the case in the NTS-NOTEARS data scenario, which also poses a challenging scenario for other algorithms. Tigramite (FullCI) achieved perfect recall across all simulations, showing a precision range of $0.51\--1.00$, indicating its tendency toward more edge detection compared to the more conservative c-GC*, with a precision range of $0.95\--1.00$. 

Parameters used for comparison algorithms: 
\begin{itemize}
    \item \textbf{NTS-NOTEARS:} \(\lambda_{1} = 0.005\), \(\lambda_{2} = 0.01\), \text{\# of lags } \(K = 2\), \(W_{thres} = 0.3\), \(h_{tol} = 1e-100\) for simulations with cycles and \(1e-8\) for those without cycles. 
    \item \textbf{TCDF:} Significance \(\alpha\) = 0.01, Learning rate = $0.01$, Epochs = $1000$, $\#$ of lags = 4 ($\tau_{max} = 4$). 
    \item \textbf{Tigramite (PCMPI \& FullCI):} \(\tau_{min} = 0.05\), \(\tau_{max} = 0.5\), \(\alpha = 0.01\), Realizations = 100 (all defaults).
\end{itemize}

\subsection{Sachs protein Dataset} \label{sec:sachs}

The Sachs protein-signalling data set is a widely used real-world benchmark for causal discovery, originally introduced by Sachs et al.~\citeyear{sachs2005causal} to reconstruct signalling pathways in primary human immune cells from multiparameter flow-cytometry measurements. The data comprise simultaneous single-cell measurements of 11 phosphorylated proteins and phospholipids collected under multiple stimulatory and inhibitory perturbation conditions, enabling evaluation of causal structure learning methods against a biologically informed consensus network. Because it combines real biological measurements, experimentally induced perturbations, and an accepted reference graph, the Sachs data set has become a standard benchmark in the causal structure learning literature, appearing in evaluations of classical Bayesian-network methods, continuous optimization approaches, neural and graph-based DAG learners, reinforcement-learning-based search procedures, and interventional causal discovery algorithms \citep{sachs2005causal, scutari2014bayesian, kalainathan2019causal, zheng2018dags, zheng2020learning, yu2019dag, goebel2020golem, lachapelle2020gradient, zhu2020causal, brouillard2020differentiable, wang2021ordering, he2021differentiable, hasan2022prior, lippe2022efficient, emezue2023benchmarking, duong2025causal}.
\begin{figure}
    \centering
    \includegraphics[width=\columnwidth]{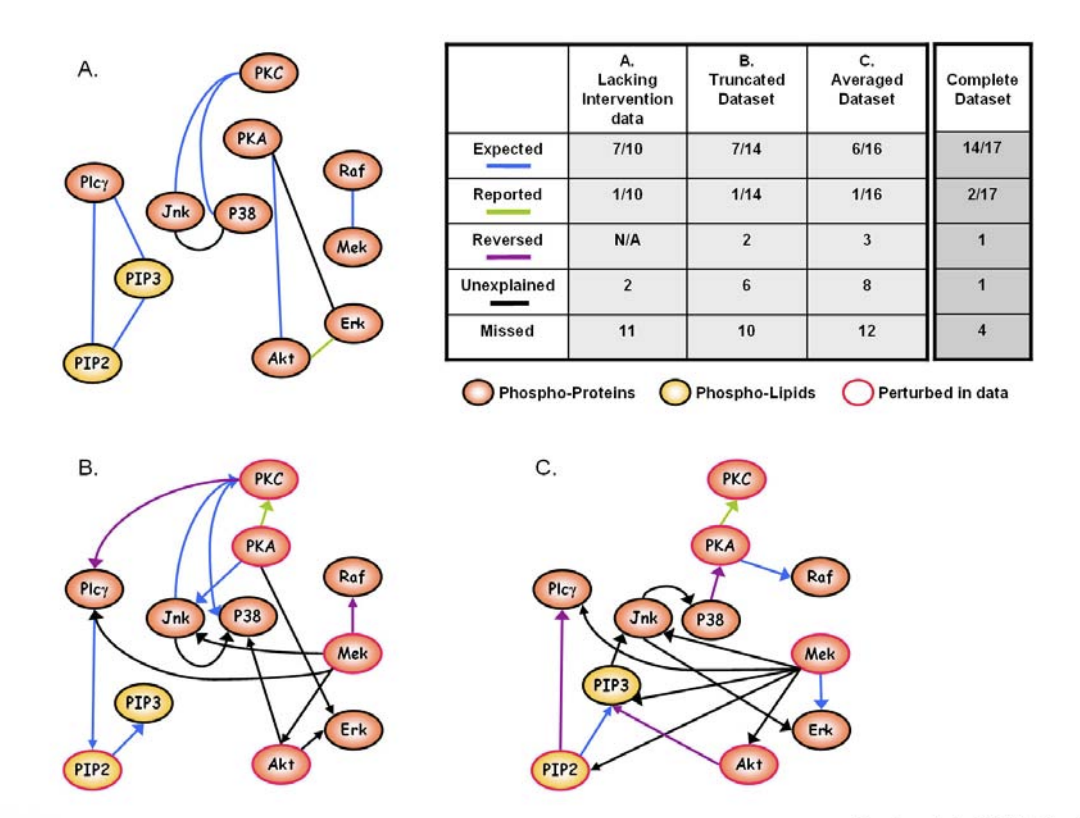}
    \caption{Ground-truth structure on interventional data: (A) Full interventional dataset (aligned with expectations), (B) Truncated dataset ($N=420$), and (C) Averaged simulated Western blot data. Adapted from \cite{sachs2005causal}.}
    \label{fig:Sachs_GT}
\end{figure}

Figure~\ref{fig:Sachs_GT} shows the ground-truth causal structure for Sachs' dataset. We inferred nearly all expected connections, missing only \emph{PKA} $\longrightarrow$ \emph{Raf}. Results varied slightly depending on the data version (Fig.~\ref{fig:Sachs_GT}a--c). While the truncated interventional dataset ($N = 420$) is suboptimal for causal learning, the third case utilised averaged simulated Western blot data (20 single-cell points per time point) to highlight the advantages of single-cell resolution.

\begin{figure}
\centering
    \includegraphics[scale=0.75]{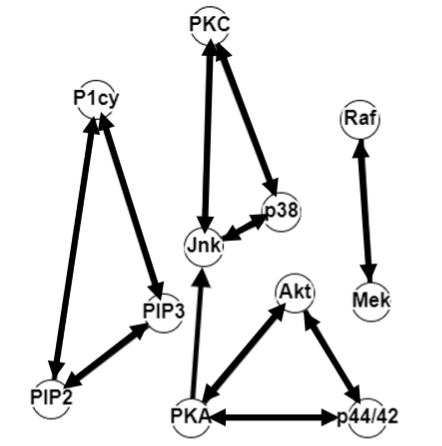}
    \caption{Aggregated \(\hat{A}\) from 14 trials of the Sachs' dataset. \emph{p44/42} is the same as the \emph{Erk}}
    \label{fig:Sachs_res}
\end{figure} 

Figure \ref{fig:Sachs_res} depicts aggregated inferences from Sachs' datasets. Both c-GC and c-GC* were consistent across trials. All expected links were inferred, including \emph{PKA} $\longrightarrow$ \emph{Jnk} uniquely inferred in Trial~11 of the truncated dataset version (Fig.~\ref{fig:Sachs_res}B). Although all expected links were inferred, they are mostly bi-directional except \emph{PKA} $\longrightarrow$ \emph{Jnk}, which was only discovered in data trial $11$. This link coincides with the expected directionality in the truncated version of the intervention data.

\subsection{Lorenz-96} \label{sec:lorenz-96}

\citeauthor{lorenz1996predictability} in \citeyear{lorenz1996predictability} introduced the Lorenz-96 model as a simplified dynamical system that combines simplicity and chaos, capturing real-world scenarios, and has led to its adoption as a benchmark dataset in CSL. The model is described by eqn \eqref{eqn:lorenz96}, which produces a sparse connectivity matrix. The parameter $F$ drives the system's chaos, while $p$ represents the number of variables.
    \begin{equation} \label{eqn:lorenz96}
        \frac{dX_k}{dt} = (X_{k+1} - X_{k-2}) X_{k-1} - X_k + F, \text{for } 1  \leq k \leq p.
    \end{equation}

We replicated the Lorenz-96 model setup in \citep{marcinkevivcs2021interpretable}, inspired by \citep{tank2021neural, karimi2010extensive}, and deployed both c-GC and c-GC* on the data. A total of $p = 40$ variables with $T = 5000$ samples were simulated. Table \ref{tab:lorenzRes} tabulates the obtained metrics. 

\begin{table}[H]
\caption{Computed metrics $(\pm \sigma)$ on Lorenz-96 dataset at various $F$. \\ $\sigma$ was computed over 10 runs}
\centering
\small
\resizebox{\columnwidth}{!}{ 
\begin{tabular}{@{}l l c c@{}}
    \hline\hline
    \textbf{F} & \textbf{Metrics} & \textbf{c-GC*} & \textbf{c-GC} \\
    \hline
    \multirow{6}{*}{10} 
    & Accuracy   & 0.9531 $\pm$ 0.0031  & 0.9506 $\pm$ 0.0027 \\
    & Precision  & 0.8058 $\pm$ 0.0021  & 0.6976 $\pm$ 0.0017 \\
    & Recall     & 0.7023 $\pm$ 0.0141  & 0.8938 $\pm$ 0.0253 \\
    & FPR        & 0.0188 $\pm$ 0.0012  & 0.0431 $\pm$ 0.0012 \\
    & SHD        & 73.646 $\pm$ 3.3532  & 74.537 $\pm$ 3.126  \\
    & BA         & 0.8406 $\pm$ 0.0035  & 0.9253 $\pm$ 0.0039 \\
    & F1         & 0.7492 $\pm$ 0.0207  & 0.7836 $\pm$ 0.0172 \\
    \midrule
    \multirow{6}{*}{40} 
    & Accuracy   & 0.9495 $\pm$ 0.0005  & 0.9497 $\pm$ 0.0008 \\
    & Precision  & 0.7426 $\pm$ 0.0033  & 0.7434 $\pm$ 0.0053 \\
    & Recall     & 0.7581 $\pm$ 0.0029  & 0.7587 $\pm$ 0.0031 \\
    & FPR        & 0.0293 $\pm$ 0.0029  & 0.0291 $\pm$ 0.0008 \\
    & SHD        & 80.023 $\pm$ 0.0123  & 80.023 $\pm$ 0.0123  \\
    & BA         & 0.8644 $\pm$ 0.0015  & 0.8648 $\pm$ 0.0016 \\
    & F1         & 0.7502 $\pm$ 0.0025  & 0.7510 $\pm$ 0.0034 \\
    \bottomrule
    \label{tab:lorenzRes}
\end{tabular}}
\end{table}

Table~\ref{tab:lorenzRes} tabulates the computed metrics on the Lorenz-96 dataset with $p = 40$, at both $F = 10$ and $40$. Both c-GC and c-GC* converged to similar metrics with $F = 40$. Slightly better results are observed at $F = 10$ compared to those observed for $F = 40$.

\section{Conclusion}

This work re-examined Granger causality through causal Bayesian networks and Reichenbach's common cause principles. The resulting view treats bivariate and multivariate GC as complementary checks: one tests temporal association, while the other asks whether that association survives conditioning on relevant histories. This interpretation led to causalised Granger causality (c-GC) and its more conservative variant, c-GC*. Across synthetic systems, common benchmarks, Sachs' protein-signalling data, and Lorenz-96 simulations, both methods recovered meaningful structure in settings with delayed effects, cycles, bidirectional links, and noisy or nonlinear dynamics. The main lesson is therefore cautious but useful: temporal prediction alone is not causation, but GC-style inference can support causal discovery when its conditioning logic is made explicit. Future work will focus on scaling c-GC to higher-dimensional systems and on sharpening its assumptions under latent confounding.

\end{NoHyper}

\printcredits
\section*{Funding statement}
\vspace{-0.8em}
This work was sponsored by Zebrafish Neuroscience Interdisciplinary Training Hub (ZENITH) under the Marie Sk{\l}odowska Curie Actions grant agreement $\#813457$.

\vspace{-0.9em}
\section*{Acknowledgements}
\vspace{-0.8em}
Sincere appreciation to Dean Rance and David Omogbhe, among other well-wishers, for their encouragement.

\vspace{-0.9em}
\bibliographystyle{cas-model2-names}

\setlength{\bibsep}{0pt}
\bibliography{cas-refs}

\end{document}